\PassOptionsToPackage{table}{xcolor} % important: 'table' option
\documentclass[10pt, a4paper]{article}

\usepackage[final]{lrec2026} % this is the new style
\usepackage{CJKutf8}
\usepackage{times}
\usepackage{latexsym}
\usepackage{tabularx}
\usepackage{tikz}

\usepackage[T1]{fontenc}
\usepackage{graphicx}
\usepackage{booktabs}
\usepackage{fontawesome5}
\usepackage{multirow}
\usepackage{amssymb}

\usepackage{pgfplots}
\pgfplotsset{compat=1.18}

\newcommand{\bay}{\texttt{bar}}
\newcommand{\deu}{\texttt{deu}}
\newcommand{\yue}{\texttt{yue}}
\newcommand{\cmn}{\texttt{cmn}}
\newcommand{\dataset}{\textsc{WiLoVa-QA}}
\newcommand{\eclektic}{ECLeKTic}

\title{Information Asymmetry Across Language Varieties: \\ A Case Study on Cantonese-Mandarin and Bavarian-German QA}

\name{\small
  \textbf{Renhao Pei}$^{\ast \textsuperscript{\faNetworkWired\kern1pt\faGraduationCap}}$, 
  \textbf{Siyao Peng}$^{\ast \textsuperscript{\faMountain\kern1pt\faRobot}}$, 
  \textbf{Verena Blaschke}$^{\textsuperscript{\faMountain\kern1pt\faRobot}}$, 
  \textbf{Robert Litschko}$^{\textsuperscript{\faMountain\kern1pt\faRobot}}$, 
  \textbf{Barbara Plank}$^{\textsuperscript{\faMountain\kern1pt\faRobot}}$
}

\address{
  \textsuperscript{\faMountain}MaiNLP, Center for Information and Language Processing, LMU Munich, Germany \\
  \textsuperscript{\faRobot}Munich Center for Machine Learning (MCML), Germany \\
   \textsuperscript{\faNetworkWired}ELLIS Institute Finland \quad  
   \textsuperscript{\faGraduationCap}University of Turku, Finland \\
  \texttt{renhaopei@gmail.com} \ \ \ \  \texttt{loganpeng1992@gmail.com}  \ \ \ \  \texttt{bplank@cis.lmu.de}
}

\abstract{
Large Language Models (LLMs) are becoming a common way for humans to seek knowledge, yet their coverage and reliability vary widely. Especially for \textit{local language varieties}, there are large asymmetries, e.g., information in local Wikipedia that is absent from the standard variant. However, little is known about how well LLMs perform under such \emph{information asymmetry}, especially on closely related languages. 
We manually construct a novel challenge question-answering (QA) dataset that captures knowledge conveyed on a local Wikipedia page, which is absent from their higher-resource counterparts---covering Mandarin Chinese vs.\ Cantonese and German vs.\ Bavarian. Our experiments show that LLMs fail to answer questions about information only in local editions of Wikipedia. 
Providing context from lead sections substantially improves performance, with further gains possible via translation.
Our topical, geographic annotations, and stratified evaluations reveal the usefulness of local Wikipedia editions as sources of both regional and global information.
These findings raise critical questions about inclusivity and cultural coverage of LLMs. 
 \\ \newline  }

\begin{document}

\maketitleabstract

\def\thefootnote{*}\footnotetext{Equal contribution.}\def\thefootnote{\arabic{footnote}}

\section{Introduction}

Large Language Models (LLMs) are becoming a commonly used channel through which humans find and learn new information  \citep{brachman2025currentfutureuselarge, elsey2025usage, YANG2025113503}.
Another information source is Wikipedia, 
% While seeking information on Wikipedia, users frequently assume that the English editions contain all knowledge \citep{hecht2010tower}. 
which includes editions in over 300 languages\footnote{\url{https://meta.wikimedia.org/wiki/List_of_Wikipedias}} and any article can be created or edited independently \citep{Lewoniewski-wikipedia-2017}, 
so that information about the same topic could vary largely across languages
% and some language editions may contain cultural or regional facts absent in other languages 
\cite{callahan2011culturalbias, bao-wikipedia-2012, wang2025wikigap}.
Till now, far less is understood about how LLMs handle facts relevant to specific cultural or linguistic communities that are less visible in global discourse \citep{myung2024blend}.

Cross-lingual retrieval for information-seeking questions remains challenging for LLMs.
% \citep{asai-choi-2021-challenges} \citep{yen-etal-2023-moqa}.
Answer retrieval during post-training remains the main bottleneck even when provided with contextual information \citep{gao-etal-2025-understanding}.
% Retrieval performances further drop sharply when the ``needle'' (relevant fact) is embedded in a long ``haystack'' (irrelevant text) with minimal lexical overlap and longer contexts \citep{modarressi2025nolima}. 
Such challenges are particularly pronounced in open-book question answering (QA), where additional external information is provided, in contrast to closed-book QA, which relies on knowledge already in the model's training \citep{yen-etal-2023-moqa}. 
LLMs also struggle to share knowledge across languages, i.e., when the question is proposed in a language different from the context language \citep{goldman2025eclekticnovelchallengeset}.

In this work, we address cross-lingual context retrieval on regional knowledge. 
Specifically, we focus on how well LLMs can answer questions drawn from knowledge in local-language Wikipedia pages but which are absent from standard-language pages. 
This is similar to the \textit{Information Asymmetry} analyzed in  \citet{roy-etal-2020-topic, roy2022information}. 
However, rather than focusing on the number of articles covered and the length of articles in English vs. non-English Wikipedia versions, we frame it as a QA task, seeking local information that is absent from the standard Wikipedia editions.
This aligns with \citet{asai-etal-2021-xor}, who observe that questions about entities from other cultures may lack answers in the questioner’s language due to cultural bias, and with the notion of a “content gap” introduced by \citet{duderstadt-2025-wikivecs}.

To address this gap, we conduct a case study of two pairs of higher-resource (\textit{standard}) versus lower-resource (\textit{local}) language varieties spoken in overlapping regions: Mandarin Chinese versus Cantonese, and German versus Bavarian. 
These pairs allow us to probe whether LLMs capture regional facts beyond what is available in dominant, higher-resource variants. 
We experiment with closed-book QA, i.e., relying on LLMs' internal parametric knowledge, and open-book QA, i.e., explicitly providing the extracted local information.
We focus on Wikipedia lead sections, which provide concise yet systematically curated knowledge that are length-friendly to test LLMs' capability in open-book QA.

We construct a new evaluation dataset, \dataset{}, consisting of facts that appear in local lead sections but are absent in their standard counterparts. Our dataset includes 294 lead section QA pairs for Cantonese–Mandarin and 178 for Bavarian–German.
Furthermore, we annotate a smaller set of document-level QA pairs for which the information is extracted from the full local Wikipedia article beyond the local lead section, yet absent from the aligned standard Wikipedia page. 
These document-level annotations introduce an additional evaluation scenario for seeking local information beyond the provided context.

Our experiments yield four key findings. First, LLMs consistently fail to answer questions derived from local Wikipedia lead sections in closed-book QA, suggesting that such knowledge is systematically underrepresented in models. 
Second, providing lead sections from local Wikipedia editions as context substantially improves performance, indicating that LLMs can reason over that knowledge when it is explicitly supplied. Translating these lead sections into the standard variant, i.e., from Cantonese to Mandarin, further boosts performance, highlighting the importance of context language for knowledge integration. 
Third, we evaluate on a cross-lingual dataset, \eclektic{} \citep{goldman2025eclekticnovelchallengeset}, and demonstrate that QA performance on local language varieties still lags behind when compared to seeking information in widely spoken languages. 
Lastly, we manually categorize \dataset{} articles to assess their topic and regional relevance and their effects on model performance.
Better performance is achieved on articles discussing more regional topics than on others.
Our results corroborate earlier findings \citep{roy-etal-2020-topic, roy2022information, wang2025wikigap} that local Wikipedia editions are not only useful as sources of regional knowledge but can also provide additional information on more region-independent topics.
% Taken together, these findings reveal both the brittleness and the potential of LLMs regarding regional knowledge.
We release our dataset and code to facilitate future research on cultural representation and coverage in LLMs.\footnote{Available at \url{https://github.com/mainlp/WiLoVa-QA}}

\section{Related Work}\label{sec:related}

% \paragraph{LLMs for Question Answering}
Earlier studies on BERT-based models have shown that these LMs encode surprising amounts of relational knowledge, facilitating open-domain QA even without fine-tuning \citep{petroni-etal-2019-language, jiang-etal-2020-know, roberts-etal-2020-much, kassner-schutze-2020-negated}.
With LLMs, information is typically accessed from model-internal parametric knowledge \citep{mallen-etal-2023-trust} acquired during pre-training, or from externally provided knowledge in the case of retrieval-augmented LLMs.
%
% Without external knowledge, prior works include, e.g., evaluations on inconsistencies in accessing parametric knowledge \citep{sedova-etal-2024-know}, answer correctness with respect to how well entities are present in pretraining data \citep{mallen-etal-2023-trust}.
LLMs degrade markedly when not prompted in English \citep{zhang-etal-2023-dont}. 
Moreover, multilingual LLMs prefer to retrieve information in the same language as the query, and when no such information exists, they rely on high-resource languages \citep{myung2024blend, sharma-etal-2025-faux}.

% \paragraph{Information Asymmetry} 
This is compounded by \textbf{information asymmetry}, i.e., not all content is equally present in volume and nature on the web \citep{roy2022information}. For example, prior work has shown that locally famous people may be described in a more favorable way in their own Wikipedia language \citep{kolbitsch2006transformation,callahan2011culturalbias}, or that some aspects are not discussed altogether in some language editions \citep{samir-etal-2024-locating,wang2025wikigap}. 
Collaborative behaviors can also vary across different language editions of Wikipedia \citep{bipat-wikipedia-2018}.
As such, it can impact the values and knowledge that LLMs learn during pretraining. 

Similarly, some \textbf{regional or local information} is often found \textbf{only} in local editions of Wikipedia. 
\citet{roy-etal-2020-topic, roy2022information} compare information coverage of English Wikipedia with eight other languages and show English articles often miss out on details in other Wikipedia editions, and that almost 50\% of non-English articles lack English correspondence.
\citet{ashrafimoghari2023informationgap} also observed such ``information gap'' between language editions. 
% that many aligned articles are not direct translations and have low semantic overlap.
This raises the question of whether LLMs reflect the same information asymmetry as on the web.

Many multilingual QA datasets are created to address this language and cultural bias, including TyDi QA \citep{clark-etal-2020-tydi}, 
XOR QA \citep{asai-etal-2021-xor},
INCLUDE \citep{romanou2025include},
BLEND \citep{myung2024blend},
and CVQA \citep{romero2025cvqa}. 
\citet{faisal-etal-2021-sd-qa} introduce a spoken dialect QA benchmark and show large performance drops for certain dialects.
% in five typologically diverse languages: Arabic, Bengali, English, Kiswahili, and Korean, 
%
Closest to our work, \citet{goldman2025eclekticnovelchallengeset} create the \eclektic{} dataset by selecting Wikipedia articles that exist \textbf{only} in one language edition and prompting Gemini to generate QA pairs. 
Model performance on \eclektic{} degrades when the question is translated and proposed to LLMs in another language. 

Our case study addresses slightly more complex situations in which Wikipedia pages exist in both local and standard language varieties, with more information on the local side. 
We investigate LLMs' asymmetric parametric knowledge gaps and their interaction with externally provided context.

\begin{table*}[t]
\centering
\scriptsize
\begin{tabularx}{\textwidth}{X}
\hline
\textbf{Lead section (Cantonese):} 
\begin{CJK}{UTF8}{gbsn}
{\CJKfamily{bsmi}
南戲係中國戲曲史上最早成熟嘅文藝戲劇，北宋末年至明朝初年（12世紀～14世紀）喺中國\textcolor{blue}{東南沿海}流行，為區別同時代嘅「北曲雜劇」，後人稱之為「南曲戲文」、「南戲」或「戲文」。南戲嘅存在，令中國嘅古代戲曲同\textcolor{cyan}{古希臘戲劇同古印度戲劇}並列為世界三大古代戲劇體系。}
\end{CJK}
\textit{Nanxi is the earliest mature literary and artistic drama in the history of Chinese drama. It was popular in the \textcolor{blue}{southeastern coast} of China from the end of the Northern Song Dynasty to the early Ming Dynasty (12th century to the 14th century). The existence of Nanxi has placed ancient Chinese drama alongside \textcolor{cyan}{ancient Greek drama and ancient Indian drama} as the three largest ancient theater systems in the world. } \\
\textbf{Lead section (Mandarin):} 
\begin{CJK}{UTF8}{gbsn}
南戏是中国戏曲史上最早成熟的戏剧类型，元朝流行于中国南方，为区别同时代的「北曲杂剧」，后人称之为南曲戏文、温州南戏或戏文。
\end{CJK}
\textit{Nanxi is the earliest mature type of drama in the history of Chinese drama. It was popular in southern China during the Yuan Dynasty. To distinguish it from the contemporary ``Beiqu Zaju'', later generations called it Nanqu Xiwen, Wenzhou Nanxi, or Xiwen.} \\
\textbf{Question 1}:
\begin{CJK}{UTF8}{gbsn}
南戏北宋末年至明朝初年（12世纪～14世纪）在中国的什么地方流行？
\end{CJK}
\textit{Where in China was Nanxi popular from the end of the Northern Song Dynasty to the beginning of the Ming Dynasty (12th to 14th century)?} 
\quad
\textbf{Answer 1}: 
\begin{CJK}{UTF8}{gbsn}
\textcolor{blue}{东南沿海。}
\end{CJK}
\textcolor{blue}{\textit{Southeast coast.}} \\
\textbf{Question 2}: \begin{CJK}{UTF8}{gbsn}
中国古代戏曲和什么并列为世界三大古代戏剧体系？
\end{CJK}
\textit{Alongside what is Chinese ancient drama regarded as one of the world’s three major ancient dramas?}
\quad
\textbf{Answer 2}: 
\begin{CJK}{UTF8}{gbsn}
\textcolor{cyan}{古希腊戏剧和古印度戏剧。}
\end{CJK}
\textcolor{cyan}{\textit{Ancient Greek drama and ancient Indian drama.}}
\\ \hline
\textbf{Lead section (Bavarian)}: 
Es Hiatamadl is a traditionella Voiksdanz, wo in Östareich und Bayern weit vabroadd is. En Danz gibts in vaschiedna Variantn. \textcolor{blue}{Da Nama kimmt vom Danzliadl ``Koa Hiatamadl mog i net''.}
\textcolor{cyan}{Da Danz hod de sejm Wuazln wia da Strohschneida, wo iwa ganz Eiropa vabroadd wor.} 
\textit{The Hiatamadl is a traditional folk dance widespread throughout Austria and Bavaria. There are various variations of the dance. \textcolor{blue}{The name comes from the dance tune ``Koa Hiatamadl mog i net (I don't like shepherd girls).''} \textcolor{cyan}{The dance has the same roots as the Strohschneider (straw cutter), which was widespread throughout Europe.}} \\
\textbf{Lead section (German)}: 
Hiatamadl ist der Name eines alpenländischen Volkstanzes, der sich in Österreich und Bayern großer Beliebtheit erfreut. \quad
\textit{Hiatamadl is the name of an Alpine folk dance that is very popular in Austria and Bavaria.} \\
\textbf{Question 1}: 
Woher kommt der Name des Tanzes Hiatamadl?
\textit{Where does the name of the dance Hiatamadl come from?}
\quad
\textbf{Answer 1}: 
\textcolor{blue}{Der Name des Tanzes kommt vom Tanzlied ``Koa Hiatamadl mog i net''. 
\textit{The name of the dance comes from the dance tune ``Koa Hiatamadl mog i net''.}
}
\\
\textbf{Question 2}: 
Aus welchem Tanz hat sich das Hiatamadl entwickelt? 
\textit{From which dance did the Hiatamadl develop?} \\
\textbf{Answer 2}: 
\textcolor{cyan}{Das Hiatamadl hat dieselben Wurzeln wie der Strohschneider, der in ganz Europa verbreitet war.
\textit{The Hiatamadl has the same roots as the Strohschneider, which was widespread throughout Europe.}} 
\\ \hline
\end{tabularx}

\caption{Sample \dataset{} annotations for \yue-\cmn{} page ``{\footnotesize \begin{CJK}{UTF8}{gbsn}南戏\end{CJK}}'' and \bay-\deu{} page ``Hiatamadl.''}
\label{tab:example}
\end{table*}

\section{\dataset{} Dataset}
\label{sec:dataset}
% Wikipedia Local Variety in Question Answering 

We create \dataset{} (\textsc{Wi}kipedia \textsc{Lo}cal \textsc{Va}riety Question Answering), a small dataset to assess LLMs' capability in retrieving information that appears on a Wikipedia page in the \textbf{local language variety} but is absent from its \textbf{standard variety} counterpart.
We discuss our target language varieties (\S\ref{subsec:target-language}), 
Wikipedia data preprocessing (\S\ref{subsec:preprocessing}), and question-answer annotations (\S\ref{subsec:annotation-guideline}).
We present detailed dataset statistics in Appendix \ref{appx:dataset-statistics}.

\subsection{Target Language Varieties}
\label{subsec:target-language}

We focus on two \textit{local} versus \textit{standard} language pairs:\footnote{For the sake of simplicity, we use the terms `local' vs.\ `standard'.
However, this terminology is somewhat flawed, as Cantonese is arguably a standardized language, especially in Hong Kong. However, speaker attitudes towards the status of Cantonese as a language (relative to Mandarin) vary greatly \cite{groves2010cantonese}.} Cantonese (ISO 639-3: \yue{}) vs.\ Mandarin Chinese (\cmn) and Bavarian  (\bay{}) vs.\ German (\deu{}).
Both pairs are spoken in overlapping regions, where the standard language prevails in formal communication, while the local variety serves as a marker of regional and cultural identity.

% Table~\ref{tab:language-dataset-stats} shows the languages and annotations.
Cantonese is a Sinitic language spoken primarily in the Guangdong province, Hong Kong, and Macau, as well as in diaspora communities outside of China. 
Bavarian is a group of non-standardized Upper German dialects spoken in southeastern Germany (Bavaria), most of Austria, and in northern Italy (South Tyrol). 
Though Cantonese and Bavarian have over 10 million speakers and are among the top 100 most spoken languages \citep{ethnologue200}, their digital presence is relatively sparse.
% \footnote{\url{https://www.ethnologue.com/insights/ethnologue200/}}
\citet{joshi-etal-2020-state} labeled Cantonese and Bavarian as ``1 - The Scraping-Bys'' (i.e., with some amount of unlabeled data) in a 0-5 scale for language resource richness (0 the rarest and 5 the richest), whereas Mandarin and German are both ``5 - The Winners'' (i.e., quintessential resource-rich languages).
For example, there are 10 times more pages on the Mandarin Chinese Wikipedia (1.5M) than on the Cantonese Wikipedia (146K), and 100 times more on the German Wikipedia (3.0M) than on the Bavarian Wikipedia (27K). 
Though underrepresented, these local pages still provide information that does not appear in standard variety pages, as shown in Table~\ref{tab:example}. 
This asymmetry makes Cantonese and Bavarian instructive test cases for studying how knowledge in regional Wikipedia editions is represented or omitted in LLMs.

\subsection{Preprocessing Wikipedia Pages}
\label{subsec:preprocessing}

\paragraph{Aligning Wikipedia pages}
We follow \citet{litschko-etal-2025-cross} to extract and align Wikipedia pages from both the local and standard varieties. 
We obtain  111.6K and 24.6K aligned Wikipedia pages for Cantonese-Mandarin (\yue{}-\cmn{}) and Bavarian-German (\bay{}-\deu{}) and covering the majority of the local pages,  76.2\% (111.6K/146.3K) of \yue{} and 90.4\% (24.6K/27.2K) of \bay{}. 

\paragraph{Local-heavy filtering}
Since our study seeks to manually annotate a small portion of information uniquely present on local pages, we further filter to local-heavy pairs. 
Namely, we only keep Wikipedia pairs for which both the \textit{lead section} (typically the introductory paragraphs)\footnote{\url{https://en.wikipedia.org/wiki/Wikipedia:Manual_of_Style/Lead_section}} and the whole document in the local language edition are longer than those of the corresponding page in the standard language. 
This resembles the \texttt{diff+} setting in \citet{roy-etal-2020-topic}, where non-English articles exceed their English counterparts in length.
We used the Wikipedia API\footnote{\url{https://pypi.org/project/Wikipedia-API/}.} to extract these lead sections (\texttt{page.summary}) and texts (\texttt{page.text}).
After filtering, only 5.38\% (6.2K/111.6K) and 4.07\% (1.0K/24.6K) of \yue{}-\cmn{} and \bay{}-\deu{} pairs remained. 

This simple filtering has some limitations. 
Our preprocessing and annotation only ensure that the information on the local-language page is absent from the \textit{paired} standard page.
It remains unknown whether such information is attested on other Wikipedia pages of the standard language, in resources outside of Wikipedia, or even in other languages beyond our study.  
Although other external sources of information may exist, our experiments and results clearly show that such information from these local Wikipedia pages remains difficult for LLMs to recall.

\paragraph{Removing direct translations and structural mismatches}
Before our annotations, we noticed that some page pairs are not ideal for our information asymmetry annotation, even after our local-heavy filtering. 
These include cases that are (i) \textit{translated}, if the local page seems to be a direct translation of the standard page (i.e., containing almost identical information and having a similar structure), regardless of whether the translated local variety sounds natural or superficial, and (ii) 
\textit{structural mismatch} if either the local page or the standard page is a disambiguation page or a page without a lead section, lacking sufficient content for our information-asymmetry annotation. 
We ask annotators to exclude these two cases from QA annotation. 
We found that more than half of inspected \bay{}-\deu{} pages, 52.5\% (261/497), are nearly direct translations, whereas only 26.6\% (61/229) for \yue{}-\cmn{} pages. 
 % (as noted in \citealt{blaschke-etal-2024-maibaam})
% are translations of each other. 
The proportion of structural mismatch pages is also higher in \bay{}-\deu{} (28.6\%) than \yue{}-\cmn{} (10.5\%). 
As a result, only 18.9\% (94/497) of \bay{}-\deu{} pairs contain distinctive content in the local page for QA annotation; comparatively, more than half of \yue{}-\cmn{} pairs, 59.8\% (137/229), deliver such information.

\subsection{Question-Answering Annotation}
\label{subsec:annotation-guideline}

We frame our local-heavy information-asymmetry annotation as a question-answering (QA) task. 
For each filtered Wikipedia pair, we ask annotators to compare both editions and identify information present only on the local page but absent from the standard. 
137 \yue-\cmn{}  and 94 \bay-\deu{} article pairs were annotated. 

We conduct QA annotations at two levels.
The main contributions are the \textit{lead section} annotations. 
Wikipedia lead sections typically range from a few dozen to a few hundred tokens (much shorter than full Wikipedia documents), making it feasible for annotators to compare pages from two language varieties and extract information present only in the local one. 
Because our study evaluates LLMs on answering such locally specific information—and compares performance with and without reference contexts in the prompt—using concise lead sections also keeps the experiments tractable.
% Since we aim to understand how well LLMs perform in answering information only present in the local Wikipedia editions, and to compare performances with or without providing reference contexts in the prompt, using concise lead sections as input contexts also makes our experiments more feasible.
We additionally annotate a smaller number of \textit{document-level} QAs to assess LLMs' capability to seek out-of-context information, which could be useful for shortcut learning.

\paragraph{Lead section QA annotations}
For each pair of lead sections, we ask annotators to identify 2-3 pieces of information that are present only on the local side and missing on the standard side. 
We follow \citet{ogundepo-etal-2023-cross} in encouraging annotators to pursue their natural curiosity in discovering information asymmetry between versions.
Annotators are instructed to (i) formulate a \textit{question} in the standard language that addresses the information, (ii) \textit{answer} the question in both the standard language and the local variety, and (iii) \textit{highlight} texts on the local section from which the answer was drawn. 
% (following \citealt{ogundepo-etal-2023-cross})
We ask annotators to create questions that (a) are as diverse as possible, (b) can be answered with a short phrase, and (c) make sense as stand-alone questions.
For example, we ask annotators to avoid frequently composing questions that ask for the year/date of an event or require multi-hop reasoning. 
% Questions sourced from a table or a figure in the lead section are also permitted. 
\dataset{} includes 294 \yue-\cmn{} and 178 \bay-\deu{} lead section QA annotations. 

\paragraph{Verifying information asymmetry in the full document}
For each lead section QA annotation, we ask annotators to verify whether the answer can be found in the full document of the standard page via a quick string search. 
If an answer is found, the supporting sentence(s) are copied into our annotation. 
 % in the full document (but not from the lead section)
Among 294 \yue-\cmn{} and 178 \bay-\deu{} lead section QA annotations, the majority, 200 (68.0\%) and 100 (56.2\%), \textbf{cannot} be answered even using the full content of the standard page.

\paragraph{Document-level QA annotations}
We supplement the \dataset{} dataset with 80 and 46 document-level QA annotations for \yue-\cmn{} and \bay{}-\deu{}. 
We ask annotators to skim through the whole Wikipedia page of the local and standard language varieties and find one piece of information from the local page that is (a) not on the standard page and (b) not already annotated at the lead section level. 
Document-level QA annotation is conducted only when some additional information from the local page can be observed within 10-15 minutes; thus, it is not annotated on every Wikipedia page pair.

\paragraph{Annotators}
We recruited two master's students in computational linguistics, one fluent in Cantonese and Mandarin and one fluent in Bavarian and German. 
Two annotators were hired and compensated at national salary rates for 50 hours of annotation per language. 
% 5 hours per week for 10 weeks between December 2024 and March 2025, 
Both annotators report that annotating each lead-section QA pair takes 5-10 minutes and that annotating each document-level QA takes 10-15 minutes, mostly due to reading the lead section and skimming both pages of the document.
Because a larger portion of \bay{}–\deu{} pages were translations or structurally mismatched, more documents had to be screened and discarded, resulting in fewer QA annotations than for \yue{}–\cmn{}.

\paragraph{Quality Control}
Since the annotators were asked to extract a few pieces of information they found prominent from paired Wikipedia pages, their QA annotations remain non-exhaustive and open-ended. 
This makes traditional overlap-based inter-annotator agreement (IAA) inappropriate to evaluate the quality of \dataset{}. 

We follow \citet{ogundepo-etal-2023-cross} and ask two project coordinators, one native Mandarin speaker and one native German, to review the elicited questions and answers that are written in Mandarin or in German.
Our quality check includes three aspects: (a) \textit{grammaticality}: if any of the \textit{question} or \textit{answer}  is ungrammatical or non-fluent; 
(b) \textit{stand-alone validity}: if a \textit{question} cannot function as an independent, context-free information-seeking question;
and (c) \textit{answer appropriateness}: if an answer does not directly and fully address its paired question when considered in isolation.
Only a few questions and answers were flagged and rewritten on the entire dataset, demonstrating the overall high quality of our original QA annotations.

\section{Experiments}\label{sec:experiment}

We evaluate state-of-the-art LLMs' ability to answer information-seeking questions whose answers are absent from the Wikipedia page in the prompting languages, namely Mandarin Chinese and German.

\begin{table*}[t]
\centering
\resizebox{\textwidth}{!}{ 
\begin{tabular}{ll|rrrrr|rrrrr}
\hline
 \multirow{2}{*}{Model} & \multirow{2}{*}{Context} & \multicolumn{5}{c|}{\yue-\cmn} & \multicolumn{5}{c}{\bay-\deu} \\
\cline{3-12}
 &  & ROUGE-L & chrF++ & BERTScore & BS-diff & LLMaJ & ROUGE-L & chrF++ & BERTScore & BS-diff & LLMaJ \\
\hline
\multirow{5}{*}{Llama3.1-8B}   & question-only
 & 13.96 & 7.81 & 17.94 & / & 7.82 
 & 11.38 & 14.29 & 11.98 & / & 15.17 
 \\
& +standard  & 14.65 & 8.20 & 19.25  & +1.31 & 9.52 
& 15.87 & 18.99 & 17.73   & +5.75 & 18.54 
\\
& +local  & \cellcolor{cyan!20} 36.63 & \cellcolor{cyan!20} 25.75 
& \cellcolor{cyan!20} 43.24   &  \cellcolor{cyan!20} +25.30 & \cellcolor{cyan!20} 69.39 
 & \cellcolor{cyan!20} \textbf{26.76} & \cellcolor{cyan!20} 27.40 & \cellcolor{cyan!20} 23.73 & \cellcolor{cyan!20} +11.75  &  \cellcolor{cyan!20} \textbf{62.36}
\\
& +standard+local & \cellcolor{cyan!20} 39.28 & \cellcolor{cyan!20} 27.64 & 43.14   & +25.20 & 65.31 
 & 26.21 & \cellcolor{cyan!20} \textbf{27.65} & \cellcolor{cyan!20} \textbf{26.28}   & 
 \cellcolor{cyan!20} \textbf{+14.30}  & 61.24 
\\
& +local (translated) &  \cellcolor{cyan!20} \textbf{47.54} & \cellcolor{cyan!20} \textbf{34.10} & \cellcolor{cyan!20} \textbf{45.94} &  
\cellcolor{cyan!20} \textbf{+28.00}  &  \cellcolor{cyan!20} \textbf{71.43} & / & / & / & /
\\
\hline
\multirow{5}{*}{Llama3.1-70B}   & question-only 
& 20.60 & 12.47 & 26.57   &  / & 14.63 &
 17.89 & 20.94 & 23.75 &  / &   24.16  \\
& +standard 
& 20.31 & 13.12 & 21.12 & \textcolor{purple}{-5.45}  &   18.37 
 & 21.04 & 23.24 & 27.15 & +3.40  &  25.28 
\\
& +local 
& \cellcolor{cyan!20}  42.20 & \cellcolor{cyan!20}  31.48 
& \cellcolor{cyan!20}  54.23 & \cellcolor{cyan!20} +27.66 &   \cellcolor{cyan!20}  \textbf{85.03} 
& \cellcolor{cyan!20} \textbf{28.77} & \cellcolor{cyan!20}  \textbf{27.33} & \cellcolor{cyan!20}  \textbf{31.13} & \cellcolor{cyan!20} \textbf{+7.38}  &  \cellcolor{cyan!20}  \textbf{75.84} 
\\
& +standard+local 
& \cellcolor{cyan!20} 45.30 & \cellcolor{cyan!20} 33.35 & 52.95  & +26.38   & 82.65 
& 28.16 & 26.57 & 30.02   & +6.27 & 73.03 \\
& +local (translated) 
& \cellcolor{cyan!20} \textbf{56.47} & \cellcolor{cyan!20} \textbf{41.81} & \cellcolor{cyan!20} \textbf{55.86} & \cellcolor{cyan!20}  \textbf{+29.29} &  82.99
& / & / & / & /
\\
\hline
\multirow{5}{*}{Qwen2.5-7B}   & question-only & 
14.40 & 7.76 & 19.17 &  / &   11.22 & 
18.53 & 25.55 & 26.32 &  /  &  12.36  \\
& +standard 
& 17.28 & 9.24 & 18.43 & \textcolor{purple}{-0.74}  &  13.61 
& 19.43 & 25.49 & 25.47 &  \textcolor{purple}{-0.85} &   11.80 
\\
& +local 
& \cellcolor{cyan!20} 49.83 & \cellcolor{cyan!20} 37.16 
& \cellcolor{cyan!20} 51.93 & \cellcolor{cyan!20} +32.76   &  \cellcolor{cyan!20} \textbf{81.97} 
& \cellcolor{cyan!20} \textbf{25.81} & \cellcolor{cyan!20} \textbf{28.88} 
& \cellcolor{cyan!20}  \textbf{30.62} & \cellcolor{cyan!20} \textbf{+4.30}  &  \cellcolor{cyan!20} \textbf{60.11} 
\\
& +standard+local 
& 48.92 & 36.31 & 50.34 & +31.17  &   78.23 
 & 24.58 & 28.31 & 30.53 & +4.21 &   56.74\\
& +local (translated) 
& \cellcolor{cyan!20} \textbf{55.42} & \cellcolor{cyan!20} \textbf{40.97} & \cellcolor{cyan!20} \textbf{53.88} & \cellcolor{cyan!20}  \textbf{+34.71}  &   80.27 
& / & / & / & /
\\
\hline
\multirow{5}{*}{Qwen2.5-72B}   & question-only & 
17.05 & 11.35 & 23.17   & / & 16.33 
& 18.56 & 23.39 & 26.96 &  / &   17.42 
\\
& +standard 
& 19.58 & 12.73 & 23.83 & +0.66  &   20.75 
 & 22.91 & 28.52 & 29.63 &  +2.67 &   20.22 \\
& +local 
& \cellcolor{cyan!20}  52.49 & \cellcolor{cyan!20}  38.79 & \cellcolor{cyan!20}  55.85 &  \cellcolor{cyan!20}  +32.68  &  \cellcolor{cyan!20}  \textbf{88.10} 
& \cellcolor{cyan!20} 35.03 & \cellcolor{cyan!20} 36.82 & \cellcolor{cyan!20} 40.28 & \cellcolor{cyan!20}  +13.32  &  \cellcolor{cyan!20} \textbf{81.46} \\
& +standard+local 
& \cellcolor{cyan!20} 53.54 & \cellcolor{cyan!20} 40.42 & 54.94 &  +31.77 
 &  85.03 
& \cellcolor{cyan!20} \textbf{37.13} & \cellcolor{cyan!20} \textbf{40.90} & \cellcolor{cyan!20} \textbf{44.65} & \cellcolor{cyan!20}  \textbf{+17.69}  &   76.97\\
& +local (translated) 
& \cellcolor{cyan!20} \textbf{58.03} & \cellcolor{cyan!20} \textbf{42.74} & \cellcolor{cyan!20} \textbf{55.92}   &  \cellcolor{cyan!20} \textbf{+32.75}  & 86.39 
& / & / & / & /
\\\hline
\multirow{5}{*}{gpt-oos-20b}   & question-only
 & 17.32 & 12.14 & 24.45   & /  & 15.65 
& 14.42 & 20.86 & 21.89   &  / & 22.47  \\
& +standard 
& 19.59 & 13.54 & 26.88 & +2.43   &  21.09 
& 18.01 & 23.85 & 24.98 & +3.09 &   25.28 \\
& +local 
& \cellcolor{cyan!20} 48.06 & \cellcolor{cyan!20} 35.60  & \cellcolor{cyan!20} \textbf{50.74} & \cellcolor{cyan!20}  \textbf{+26.29}  &  \cellcolor{cyan!20} \textbf{88.44} 
& \cellcolor{cyan!20} \textbf{26.33} & \cellcolor{cyan!20} \textbf{29.30} & \cellcolor{cyan!20} \textbf{31.04} &  \cellcolor{cyan!20} \textbf{+9.15} &  \cellcolor{cyan!20} \textbf{70.79}  
\\
& +standard+local 
& 47.15 & 35.07 & 48.29 & +23.84 
 &   86.39 
& 25.14 & 27.84 & 29.59 & +7.70  &  68.54  \\
& +local (translated) 
& \cellcolor{cyan!20} \textbf{50.69} & \cellcolor{cyan!20} \textbf{37.15} & 50.56   & +26.11 & 85.37 
& / & / & / & /
\\\hline
\multirow{5}{*}{gpt-oos-120b}   & question-only
& 20.91 & 14.41 & 26.24 &  / &   20.41
& 17.75 & 23.85 & 25.98 &  /  &  23.03 
\\
& +standard 
& 23.58 & 16.18 & 27.98 & +1.74   &  32.65 
& 20.33 & 26.35 & 28.06 & +2.08 &   26.97 
\\
& +local 
& \cellcolor{cyan!20} 48.34 & \cellcolor{cyan!20} 35.56 & \cellcolor{cyan!20} 50.13 & \cellcolor{cyan!20}  +23.89  &  \cellcolor{cyan!20} \textbf{89.46} 
& \cellcolor{cyan!20} \textbf{28.90} & \cellcolor{cyan!20} 32.41 & \cellcolor{cyan!20} 34.81 &  \cellcolor{cyan!20} +8.83  &  \cellcolor{cyan!20} \textbf{73.60} 
\\
& +standard+local 
& 48.21 & 35.34 & 49.01   & +22.77  & 86.39 
& 28.69 & \cellcolor{cyan!20} \textbf{33.44} & \cellcolor{cyan!20} \textbf{35.03}   & \cellcolor{cyan!20} \textbf{+9.05} & 73.03 
\\
& +local (translated) 
 &\cellcolor{cyan!20} \textbf{51.33} & \cellcolor{cyan!20} \textbf{37.73} & \cellcolor{cyan!20} \textbf{51.12} & \cellcolor{cyan!20} \textbf{+24.88}  &  86.39 
& / & / & /   &  / & /
\\
\hline
\end{tabular}
}
\caption{Lead section QA performance, where the key information required to answer the questions is contained in the \textit{+local} context. \textbf{Bold} indicates the highest score and \textcolor{cyan!50}{Blue} highlights the winner between \textit{+local} and \textit{+standard}, and all other settings higher than the winner. BS-diff measures the difference in BERTScore compared to the \textit{question-only} baseline for the same LLM and prompt scenario.}
\label{tab:results}
\end{table*}

\paragraph{Models}
We evaluate six LLMs from three families: Llama3.1-8B/70B \citep{grattafiori2024llama3herdmodels}, Qwen2.5-7B/72B \citep{qwen2025qwen25technicalreport}, and gpt-oss-20B/120B \citep{openai2025gptoss120bgptoss20bmodel}.
We use vLLM \citep{kwon2023efficient} as the inference backend for efficient LLM decoding. 
We set a low temperature of 0.3 to encourage more deterministic and consistent responses for the QA task.
Models automatically choose the compute precision based on their respective settings (\texttt{dtype=`auto'}), with a maximum context length of 10K tokens and a maximum output length of 5K tokens.

\paragraph{Datasets}
We use our lead section and document QA annotations to evaluate LLMs' ability to answer information-seeking questions. 
We also evaluate on \eclektic{} \citelanguageresource{goldman2025eclekticnovelchallengeset} to compare performance when retrieving information from another standard language (e.g., French) versus a local variety (i.e., Cantonese or Bavarian).

Unlike our manual QA annotations, the \eclektic{} dataset generated knowledge-seeking questions in 12 source languages using Gemini \citep{geminiteam2025geminifamilyhighlycapable}, for which the answer appears in a relevant Wikipedia article in the source language, but the corresponding Wikipedia pages in the other 11 languages are non-existent. 
Questions and answers are translated into the other 11 languages to evaluate LLMs in cross-lingual knowledge transfer. 
To compare with our \dataset{} results, we assess our LLM selection on a subset of 333 items sourced from the 10 languages, excluding Mandarin Chinese (35 items) and German (16 items). 
It is also worth noting that the extracted context in \eclektic{} includes the first 10 sentences of the Wikipedia pages, which differ slightly from the lead sections in \dataset{}.
Additionally, the answers in \eclektic{} are much shorter (on average 8.3 characters for \cmn{} and 21.0 for \deu{}) than \dataset{} (21.7 for \cmn{} and 83.9 for \deu{}).

\paragraph{Context scenarios}
We prompt LLMs to answer information-seeking questions in \dataset{} when given different context scenarios:
(1a) \textit{question-only}: asking the question without any additional context information; 
(1b) \textit{+standard}: providing lead section text from the  Wikipedia article in the standard language as context; 
(1c) \textit{+local}: providing lead section text from the  Wikipedia article in the local language;
(1d) \textit{+standard+local}: providing lead section texts from both the standard and the local from Wikipedia articles. 
Since the QA annotations are sourced from the Wikipedia pages in the local languages, \textbf{we expect \textit{+local} and \textit{+local+standard} to achieve better performance} than the \textit{question-only} baseline and \textit{+standard}.
To isolate retrieval deficiencies attributable to cross-lingual retrieval rather than context information, we additionally assess (1e) \textit{+local (translated)} on \yue-\cmn{} items: providing translated lead section (\yue$\rightarrow$\cmn{}) by using Google Cloud Translation.\footnote{\url{https://cloud.google.com/translate}} 
However, due to the lack of existing high-quality sentence-level Bavarian-to-German translators (further discussed in Appendix~\ref{appx: bar-translation-challenge}), we leave it to future work to evaluate the \textit{+local (translated)} context for \bay-\deu{}.

For \eclektic{}, we used their translations of questions and answers into Mandarin/German and similarly tested three context scenarios: 
(2a) \textit{question-only}: without any additional context information; 
(2c) \textit{+source}: providing context text from the source Wikipedia article in another language; 
and (2e) \textit{+source (translated)}: providing the translated context text (others$\rightarrow$\cmn{}/\deu{}).
As \eclektic{} addresses Wikipedia pages absent in the target languages, it lacks counterparts to (1b) and (1d).

Due to concerns on prompt-sensitive knowledge (i.e., facts retrieved only under specific prompt templates, \citealt{yin-etal-2024-benchmarking}), we experiment with three prompt variations in our QA experiments. However, no significant performance difference is observed, and, for consistency, we report results using one of the prompt templates (see Appendix~\ref{appx:prompts}).

\paragraph{Metrics}
We use standard evaluation metrics for Natural Language Generation (NLG) tasks to measure the similarity between the predicted and reference answers at the lexical, character, and semantic levels.
ROUGE-L \citep{lin-2004-rouge} evaluates lexical overlap of the longest common subsequence, chrF++ \citep{popovic-2015-chrf}  captures character-level similarity that is robust across languages and morphological variants, and BERTScore \citep{Zhang2020BERTScore} measures token-level semantic similarity using contextualized embeddings from pretrained language models. We report F1 for ROUGE-L and BERTScore.

To complement traditional NLG evaluation metrics, we also employ LLM-as-a-judge (LLMaJ, \citealt{chiang-lee-2023-large, NEURIPS2023_91f18a12}). 
The judge model (gpt-oss-20b) takes the question, the reference answer, and the generated answer as input and determines whether the generated answer is correct.
We adapt the evaluation prompt from \citet{goldman2025eclekticnovelchallengeset}, which is provided in Appendix~\ref{appx:prompts}.
To verify the reliability and consistency of LLMaJ as an evaluation method, we manually inspected 50 QA instances evaluated by LLMaJ and found only one misjudgment (see Appendix~\ref{appx: error analysis}).

\paragraph{Lead section results}
Table~\ref{tab:results} presents our main results on lead section QA. 
Adding a standard lead section (\textit{+standard}) as context only shows marginal gain (e.g., +3.40 on \bay-\deu{} Llama3.1-70B) or even sometimes hurts performance (e.g., -5.45 on \yue{}-\cmn{} Llama3.1-70B), where \textit{+local} substantially improves QA performance over the baseline (e.g., +27.66 on \yue{}-\cmn{} and +7.38 on  \bay-\deu{} Llama3.1-70B). 
This agrees with our hypothesis that providing relevant context in the local language enhances the quality of LLM-generated answers.
\textit{+standard+local} generally achieves similar performance to \textit{+local}, without further gains.

\begin{table}[t!bh]
\centering
\scriptsize
\begin{tabularx}{0.5\textwidth}{X}
\hline
\textbf{Question}:
Woher kommt der Name des Tanzes Hiatamadl? \textit{Where does the name of the dance Hiatamadl come from?} \\
\textbf{Reference Answer}: 
Der Name des Tanzes kommt vom Tanzlied ``Koa Hiatamadl mog i net''. 
\textit{The name of the dance comes from the dance tune ``Koa Hiatamadl mog i net''.} \\
\hline
\textbf{\textit{question-only} answer:}
Der Name des Tanzes Hiatamadl \textcolor{purple}{stammt aus der Mundart des Wiener Umlands} und bedeutet so viel wie \textcolor{purple}{„hier tanzt man“}.  \textit{The name of the dance Hiatamadl \textcolor{purple}{comes from the dialect of the Viennese surroundings} and means something like \textcolor{purple}{``here we dance''}.}\\
\hline
\textbf{\textit{+standard} answer:}
Der Name ``Hiatamadl'' stammt aus dem Bayerischen und setzt sich zusammen aus \textcolor{purple}{``Hiat'' (was ``Hügel'' oder ``Anhöhe'' bedeutet)} und ``Madl'' (was ``Mädchen'' bedeutet). Der Tanz wird also als \textcolor{purple}{``Tanz der Hügelmädchen'' interpretiert, was auf die alpenländische Herkunft und die traditionelle Aufführung durch junge Frauen hinweist}. 
\textit{The name ``Hiatamadl'' comes from Bavarian and is composed of \textcolor{purple}{``Hiat'' (meaning ``hill'' or ``elevation'')} and ``Madl'' (meaning ``girl''). The dance is therefore interpreted as \textcolor{purple}{``Dance of the Hill Girls,'' which refers to its Alpine origins and the traditional performance by young women}.} \\
\hline
\textbf{\textit{+local} answer:}
Der Name des Tanzes Hiatamadl kommt vom Lied ``Koa Hiatamadl mog i net''. \textit{The name of the dance Hiatamadl comes from the song ``Koa Hiatamadl mog i net''.} \\
\hline
\textbf{\textit{+standard+local} answer:}
Der Name des Tanzes Hiatamadl kommt vom gleichnamigen Lied ``Koa Hiatamadl mog i net''. \textit{The name of the dance Hiatamadl comes from the song of the same name ``Koa Hiatamadl mog i net''.}\\
\hline
\end{tabularx}
\caption{Sample Qwen2.5-72B answers. \textit{+local} and \textit{+standard+local} answers are alike the reference, whereas \textit{question-only} and \textit{+standard} answers contain misinformation highlighted in \textcolor{purple}{purple}.}
\label{tab:example-llm-output}
\end{table}

The improvements on \yue{}-\cmn{} are consistently larger than those observed on \bay{}-\deu{}. 
We hypothesize this might be due to LLMs' poorer ability to extract relevant information from Bavarian, a lower-resource language than Cantonese. 
Table \ref{tab:example-llm-output} illustrates an example of the generated answers by our best-performing model Qwen2.5-72B on the \bay{}-\deu{} example in Table~\ref{tab:example}.
\textit{+question-only} and \textit{+standard} answers contain misinformation, including falsely claiming that the dance comes from ``Viennese surroundings'' and misinterpreting ``Hiata'' as hill, which means shepherd in Bavarian (More examples can be found in Appendix~\ref{appx: error analysis}).

Moreover, when local-to-standard translation is available, for \yue{}$\rightarrow$\cmn{}, \textit{+local (translated)} tends to further improve scores on ROUGE-L, chrF++, and BERTScore, achieving the highest score. 
However, when measured by the accuracy of LLM-as-a-judge, \textit{+local} achieves the highest performance, except Llama3.1-8B, where \textit{+local (translated)} performs the best.
This discrepancy may stem from the different nature of the metrics: LLMaJ directly assesses answer correctness, whereas the overlap-based metrics measure similarity with the reference text. 
These two evaluation perspectives are complementary, and we leave it to future work to systematically align these metrics with human judgments. 
Moreover, the logographic nature of Chinese characters (simplified for Mandarin and traditional for Cantonese) may lead to inconsistent evaluations between the two language groups at the character and token levels (ROUGE-L and chrF++). 
Therefore, we use the semantically based BERTScore for the follow-up results and analyses and will present other metrics in Appendix~\ref{appx:complete results}.

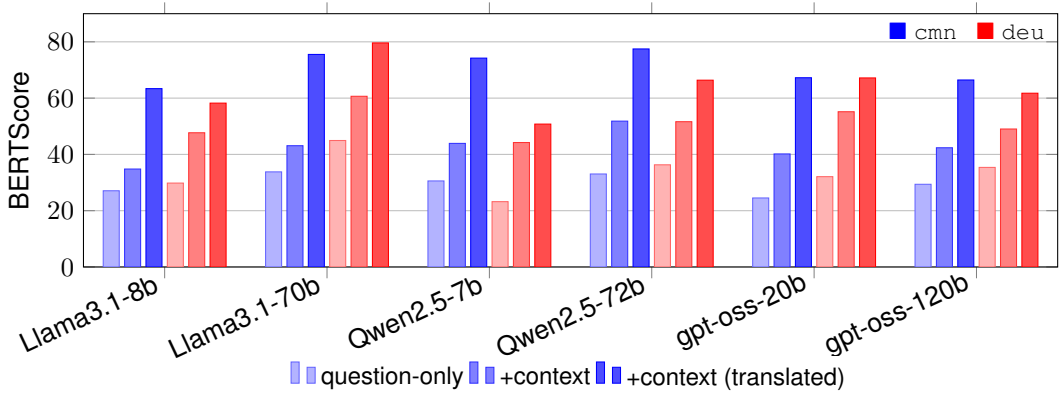
\begin{figure*}[t]
\centering
\begin{tikzpicture}
\begin{axis}[
  width=0.9\textwidth,
  height=0.2\textheight,
  ybar,
  bar width=6pt,
  enlarge x limits=0.1,
  ylabel={BERTScore},
  ymin=0, ymax=90,
  ymajorgrids,
  xtick=data,
  xticklabel style={rotate=25, anchor=east, align=center},
  symbolic x coords={Llama3.1-8b,Llama3.1-70b,Qwen2.5-7b,Qwen2.5-72b,gpt-oss-20b,gpt-oss-120b},
  legend style={at={(0.5,-0.35)},anchor=north,legend columns=3,draw=none,font=\small}
]

% Mandarin results - shades of blue
\addplot[color=blue!60, fill=blue!30] coordinates {(Llama3.1-8b,27.08) (Llama3.1-70b,33.79) (Qwen2.5-7b,30.57) (Qwen2.5-72b,33.04) (gpt-oss-20b,24.52) (gpt-oss-120b,29.38)};
\addplot[color=blue!80, fill=blue!50] coordinates {(Llama3.1-8b,34.80) (Llama3.1-70b,43.09) (Qwen2.5-7b,43.91) (Qwen2.5-72b,51.82) (gpt-oss-20b,40.16) (gpt-oss-120b,42.36)};
\addplot[color=blue!100, fill=blue!70] coordinates {(Llama3.1-8b,63.37) (Llama3.1-70b,75.52) (Qwen2.5-7b,74.21) (Qwen2.5-72b,77.47) (gpt-oss-20b,67.24) (gpt-oss-120b,66.44)};

% German results - shades of red
\addplot[color=red!60, fill=red!30] coordinates {(Llama3.1-8b,29.81) (Llama3.1-70b,44.94) (Qwen2.5-7b,23.21) (Qwen2.5-72b,36.30) (gpt-oss-20b,32.09) (gpt-oss-120b,35.38)};
\addplot[color=red!80, fill=red!50] coordinates {(Llama3.1-8b,47.69) (Llama3.1-70b,60.65) (Qwen2.5-7b,44.22) (Qwen2.5-72b,51.62) (gpt-oss-20b,55.16) (gpt-oss-120b,49.02)};
\addplot[color=red!100, fill=red!70] coordinates {(Llama3.1-8b,58.22) (Llama3.1-70b,79.61) (Qwen2.5-7b,50.77) (Qwen2.5-72b,66.37) (gpt-oss-20b,67.17) (gpt-oss-120b,61.74)};

\legend{question-only,+context,+context (translated)}
\end{axis}

% Add color coding explanation
\node[anchor=north west] at (rel axis cs:0.9, 1.0) {\small \textcolor{blue}{$\blacksquare$} \cmn{}};
\node[anchor=north west] at (rel axis cs:0.99, 1.0) {\small \textcolor{red}{$\blacksquare$} \deu{}};
\end{tikzpicture}
\caption{QA performance measured by BERTScore on the ECLeKTic dataset. Results compare context types of \textit{question-only}, \textit{+context}, and \textit{+context (translated)} across models. 
}
\label{fig:bertscore_grouped_eclektic}
\end{figure*}

\begin{table}[t]
\centering
\resizebox{0.5\textwidth}{!}{ 
\begin{tabular}{ll|r|r}
\hline
 \multirow{1}{*}{Model} & \multirow{1}{*}{Context} & \yue-\cmn & \bay-\deu \\
\hline
\multirow{5}{*}{Llama3.1-8B}   
& question-only & 22.42 & 11.61 \\
& +standard & +1.55 & +1.18 \\
& +local & -0.24 & -0.40 \\
& +standard+local  & \textbf{+5.06} & \textbf{+7.99} \\
& +local (translated)  & +1.06 & / \\
\hline
\multirow{5}{*}{Llama3.1-70B}   
& question-only  & 24.48 & 22.39 \\
& +standard & -0.49 & \textbf{+8.61} \\
& +local & +0.42 & +1.63 \\
& +standard+local & +0.68 & +5.59 \\
& +local (translated) & \textbf{+1.62} & / \\
\hline
\multirow{5}{*}{Qwen2.5-7B}   
& question-only  & 23.09 & 27.39 \\
& +standard & +0.80 & \textbf{+2.43} \\
& +local & -1.08 & -5.17 \\
& +standard+local & \textbf{+1.01} & -6.12 \\
& +local (translated) & -4.36 & / \\
\hline
\multirow{5}{*}{Qwen2.5-72B}   
& question-only  & 25.83 & 25.36 \\
& +standard & +2.00 & +4.02 \\
& +local & \textbf{+3.25} & \textbf{+7.54} \\
& +standard+local & +1.55 & +6.64 \\
& +local (translated) & +0.19 & / \\
\hline
\multirow{5}{*}{gpt-oss-20b}   
& question-only  & 22.44 & 22.59 \\
& +standard & -1.13 & +3.95 \\
& +local & +0.72 & \textbf{+6.32} \\
& +standard+local & +1.24 & +3.09 \\
& +local (translated) & \textbf{+3.59} & / \\
\hline
\multirow{5}{*}{gpt-oss-120b}   
& question-only  & 23.23 & 23.38 \\
& +standard & +3.05 & +6.15 \\
& +local & +1.72 & +5.75 \\
& +standard+local & \textbf{+3.95} & \textbf{+7.51} \\
& +local (translated) & +1.68 & / \\
\hline
\end{tabular}
}
\caption{
Document-level BERTScore difference from \textit{question-only} baseline; \textbf{bold} indicates best.
}
\label{tab:results2-bertscore-diff-bold}
\end{table}

\paragraph{Document-level results}
Table~\ref{tab:results2-bertscore-diff-bold} presents results on document-level QA annotations to evaluate LLMs' capability in answering out-of-context questions. 
Adding additional context, such as \textit{+standard}, \textit{+local}, or \textit{+local (translated)}, slightly improves performance over the \textit{question-only} baseline.
This minimal improvement is expected, as the added contexts lack the crucial information needed to answer the questions. 
Nevertheless, even context without answer-relevant information may provide limited benefits by helping the LLM better identify the topic or domain of the question. 
Overall, our results show that answer-relevant context is key to LLM QA performance.

\paragraph{\eclektic{} cross-lingual results}
We additionally report \eclektic{}'s cross-lingual retrieval results in 
Figure~\ref{fig:bertscore_grouped_eclektic}.
A clear and consistent pattern emerges: performance improves noticeably when the additional context contains the key information, even when the context language differs from the prompt language. 
The gains are even larger when the context is translated into the prompt language, consistent with the trend observed in the results of \dataset{}'s lead section.
Because languages in the \eclektic{} dataset are relatively high-resource, performance on \dataset{} (Table \ref{tab:results}) is lower than on \eclektic{} (Figure \ref{fig:bertscore_grouped_eclektic}), indicating that our low-resource varieties, Cantonese and Bavarian, pose greater challenges for LLMs.

\section{Article Region and Topic}
\label{sec:analysis}

This section provides a fine-grained analysis of \dataset{}'s \textbf{geographical and topic diversity}.
We conduct a stratified evaluation using these labels and investigate whether QA performance is more challenging for specific regions or topics. 

\paragraph{Topic classification}
To characterize the topical coverage of our dataset, we assign a topic label to each selected article.
Two leaders of the project (one fluent in Mandarin and one fluent in German) independently annotated the relevant language partition, and after adjudication, 14 topic categories were found relevant: 
\textit{history, sport, geography, entertainment-art, animals-plants, politics-government, food, transportation, science-math-technology, linguistics, culture-customs, education, business}, and \textit{other}.

Among the categories, \textit{geography} (26 \yue-\cmn{} and 35 \bay-\deu{} articles) and \textit{entertainment-art} (25 \yue-\cmn{} and 19 \bay-\deu{}) are the two most frequent in both languages. 
In our \bay-\deu{} data, \textit{geography} includes different towns in Bavaria and Austria, but also outside of the Bavarian speaker regions, such as in Poland, the Faroe Islands, China, etc. 
\textit{Entertainment-art} includes musicians, poets, movies, both in the local region and beyond. 
For \yue-\cmn{}, 
\textit{geography} is also well balanced between street names and towns in the Cantonese-speaking regions, and location entities that are further away, such as in Spain, Australia, etc. 
\textit{Entertainment} includes many TV shows and movies produced in Guangdong, Hong Kong, or Macau.

\paragraph{Region classification}
We also annotate each article for region relevance to understand the proportion of \yue{} and \bay{} articles that discuss concepts and cultures in the Cantonese- and Bavarian-speaking regions. 
Based on lead sections and Wikipedia categories, we annotate if the articles discuss information pertinent to the region where \textit{local} variety is spoken, general to the \textit{standard} language region, or \textit{others}.
\textit{Local} includes regions such as Guangdong, Guangxi, Hong Kong, and Macau for \yue{} and Bavaria, Austria, South Tyrol for \bay{}.
The \textit{standard} region includes Greater China and Taiwan, as well as the rest of Germany, Italy, and Switzerland. 
Additionally, we define an \textit{others} category to include articles pertinent to other specific regions, e.g., 16 \yue{}-\cmn{} articles discuss Japanese soccer players, as well as concepts that are region-independent, such as ``triangle ruler,'' ``time,'' or ``Doppler effect.''
Surprisingly, in both \yue{}-\cmn{} and \bay-\deu{} samples, less than half are \textit{local} articles: 49 (35.8\%) and 35 (37.2\%).
The majority are \textit{other} articles that discuss concepts beyond the local or standard language region.

\begin{table}[t]
\centering
\resizebox{0.5\textwidth}{!}{
\begin{tabular}{ll|r|r}
\hline
Topic/Region & Context & \yue-\cmn & \bay-\deu \\
\hline
\multirow{4}{*}{\textit{geography}}
& question-only & 25.16 & 28.17 \\
& +standard & 22.99 & 31.08 \\
& +local & \textbf{57.02} & 37.23 \\
& +standard+local & 56.34 & \textbf{43.67} \\
\hline
\multirow{4}{*}{\textit{entertainment-art}}
& question-only & 24.00 & 22.79 \\
& +standard & 28.26 & 27.59 \\
& +local & 56.46 & 40.42 \\
& +standard+local & \textbf{57.18} & \textbf{44.07} \\
\hline
\multirow{4}{*}{\textit{local region}} 
& question-only & 26.04 & 26.86 \\
& +standard & 26.26 & 31.65 \\
& +local & \textbf{59.55} & 43.46 \\
& +standard+local & 58.57 & \textbf{49.19} \\
\hline
\multirow{4}{*}{\textit{all articles (avg)}}   & question-only & 
 23.17 & 
 26.96 
\\
& +standard 
& 23.83 
& 29.63   \\
& +local 
&   \textbf{55.85} 
& 40.28  \\
& +standard+local 
& 54.94  
&  \textbf{44.65}  \\
\end{tabular}
}
\caption{Lead section BERTScore results of Qwen2.5-72B on top 2 topics and \textit{local} articles.}
\label{tab:qwen72b_topic_region_bertscore-short}
\end{table}

\paragraph{Results}
Table \ref{tab:qwen72b_topic_region_bertscore-short} presents Qwen2.5-72B (overall best performing model) results on the top 2 topic labels and \textit{local}-region articles (full results on all topic and region labels are in Appendix~\ref{appx:all tpoic and region results}). 
In general, these stratified performances match the dataset's average, indicating that \textit{local}-relevant articles are \textbf{not} more challenging for LLMs than other articles. 
On \yue{}-\cmn{}, we observe that explicitly inserting lead sections from Cantonese achieves the best (or near best for \textit{entertainment-art}) performances.
A recent evaluation benchmark, HKCanto-Eval \citep{cheng-etal-2025-hkcanto}, also shows that Qwen2-72B \citep{yang2024qwen2technicalreport} outperforms most models on academic, professional, and cultural QA for  Cantonese.
Overall, our \dataset{} data provides a good testbed for low-resource evaluation scenarios of information asymmetry within and beyond regional knowledge.

\section{Conclusion}\label{sec:conclusion}

This paper introduces \dataset{}, a Wikipedia-based question-answering dataset that addresses information asymmetry across language varieties. 
Through experiments on Cantonese–Mandarin and Bavarian–German, we reveal that current LLMs often fail to recall local knowledge in closed-book settings, but can retrieve it effectively once the relevant local context is provided—highlighting a clear information asymmetry between high- and low-resource varieties.
The findings suggest that knowledge gaps in LLMs parallel those observed on the web, where regional content remains underrepresented. 
Beyond QA, \dataset{} offers a foundation for studying cross-variety understanding, cultural bias, and regional knowledge access in multilingual LLMs.

\section*{Acknowledgements}

We thank Miriam Winkler and Yuchen Mao for their contributions and invaluable feedback on \dataset{}.
This work is funded by the ERC Consolidator Grant DIALECT 101043235.

% \newpage

%%%%%%%%%%%%%%%%%%%%%%%%%%%%%%%%%%%%%%%%%%%%%%%
%%%%%%%%%%%%%%%% Bib below %%%%%%%%%%%%%%%%%%%%
%%%%%%%%%%%%%%%%%%%%%%%%%%%%%%%%%%%%%%%%%%%%%%%

\section{References}\label{sec:reference}
\bibliographystyle{lrec2026-natbib}
\bibliography{dialect-qa}

\section{Language Resource References}
\label{lr:ref}
\bibliographystylelanguageresource{lrec2026-natbib}
\bibliographylanguageresource{languageresource}

\appendix

\section{\dataset{} statistics} 
\label{appx:dataset-statistics}

\begin{table}[h]
\centering
\resizebox{0.48\textwidth}{!}{
\begin{tabular}{c|cc|cc}
\textit{language} & Bavarian & German & Cantonese & Mandarin \\
\hline
\textit{ISO 639-3} 
& \bay & \deu & \yue & \cmn
\\
\begin{tabular}[c]{@{}c@{}} \textit{\#speakers}  \\ \textit{(\#rank)}\footnote{According to The Ethnologue 200 \citep{ethnologue200} \url{https://www.ethnologue.com/insights/ethnologue200/}.}
\end{tabular} 
& \begin{tabular}[c]{@{}c@{}} 13.7M \\ (\#92) \end{tabular}
& \begin{tabular}[c]{@{}c@{}} 134.0M \\ (\#12) \end{tabular}
& \begin{tabular}[c]{@{}c@{}} 85.7M \\ (\#25) \end{tabular}
& \begin{tabular}[c]{@{}c@{}} 1.2B \\ (\#2) \end{tabular}
 \\
\textit{\#wiki-pages\footnote{As of September 23, 2025.}} 
& 27.2K & 3,053.0K &  146.3K & 1,501.5K \\
\hline
\textit{\#aligned-pages} &   \multicolumn{2}{c|}{24.6K} &   \multicolumn{2}{c}{111.6K}  \\
\textit{\#local-heavy-pages} &   \multicolumn{2}{c|}{1,038} &   \multicolumn{2}{c}{6,173}  \\
\hline
\textit{\#inspected-pages} &  \multicolumn{2}{c|}{497} &  \multicolumn{2}{c}{229} \\
\textit{\#qa-pages}  &  \multicolumn{2}{c|}{94} & \multicolumn{2}{c}{137} \\
\textit{\#lead section-qa} &  \multicolumn{2}{c|}{178} &  \multicolumn{2}{c}{294} \\
\textit{\#doc-qa} & \multicolumn{2}{c|}{46} & \multicolumn{2}{c}{80} \\
\end{tabular}
}
\caption{Statistics of the \dataset{} dataset and included language varieties.}
\label{tab:language-dataset-stats}
\end{table}

\section{Challenges in Bavarian–German translation}
\label{appx: bar-translation-challenge}

Machine translation involving Bavarian remains largely underexplored. Prior work mainly relies on LLMs to translate from Bavarian to German \citep{litschko-etal-2025-make,hoffmann2025llamagenba10btrilinguallargelanguage}, and to the best of our knowledge, the only study training a dedicated MT system for Bavarian is \citet{her-kruschwitz-2024-investigating}. In contrast, high-performing translation systems exist for Cantonese. To avoid confounding our question-answering results with translation errors, we conduct translation experiments only on Cantonese.

\section{Prompt Templates}
\label{appx:prompts}

\begin{itemize}

\item \textbf{Prompt template: baseline}\\
Please answer this question:
\{question\}\\
Please enclose your final answer in <Answer>...</Answer> tags.
Please make sure to answer in English.

\item \textbf{Prompt template: standard language summary}\\
Please answer this question:
\{question\}\\
Here is some relevant information:
\{standard language summary\}\\
Please enclose your final answer in <Answer>...</Answer> tags.
Please make sure to answer in English.

\item \textbf{Prompt template: dialect summary}\\
Please answer this question:
\{question\}\\
Here is some relevant information:
\{dialect summary\}\\
Please enclose your final answer in <Answer>...</Answer> tags.
Please make sure to answer in English.

\item \textbf{Prompt template: standard language summary + dialect summary}\\
Please answer this question:
\{question\}\\
Here is some relevant information:
\{standard language summary\}\\
Here is some relevant information:
\{dialect summary\}\\
Please enclose your final answer in <Answer>...</Answer> tags.
Please make sure to answer in English.

\item \textbf{Prompt template of LLM-as-a-judge}\\
Your task is to evaluate whether a generated answer correctly answers the question, using the provided ground truth answer as reference.\\
Question:
\{question\}\\
Ground Truth Answer:
\{ground truth\}\\
Generated Answer:
\{generated answer\}\\
Please output a single capitalized word (YES or NO) as evaluation result, without any additional explanation:\\
    - YES: the generated answer is correct according to the ground truth answer.\\
    - NO: the generated answer is incorrect according to the ground truth answer.

\end{itemize}

\section{Error examples}
\label{appx: error analysis}

\begin{CJK*}{UTF8}{gbsn}

\begin{itemize}
\item \textbf{LLMaJ Misjudgment}\\
index: 204\\
question: 肥冢一晃曾效力过哪些俱乐部？ \textit{(Which clubs has Kazuaki Koezuka played for?)}\\
answer: 大阪飞脚，京都不死鸟 \textit{(Gamba Osaka, Kyoto Sanga)}\\
ground truth answer: 大坂飞脚和京都不死鸟。 \textit{(Gamba Osaka and Kyoto Sanga.)}\\
LLMaJ: NO\\

\item \textbf{QA error even when provided with sufficient context (gpt-oss-120B)}\\
index: \textit{170}\\
question: 《五个小孩的大叔》的节目主持是谁？ 
\textit{(Who is the host of the program "5 Kids and a Bloke"?) }\\
prompt: 请回答以下问题：《五个小孩的大叔》的节目主持是谁？这里有一些相关的信息：《五个小孩的大叔》（英语：5 Kids And A Bloke，粤文意思：五个孩子的大叔）是香港电视广播有限公司制作的真人骚节目，全节目共四集，在2016年10月2号到23号逢礼拜日晚20:00-20:30在翡翠台播，并在myTV提供节目重温。节目主持是洪永城和电影《五个小孩的校长》里五个小主角：何涴瀠、傅舜盈、王诗雅、李咏珊、陈丽儿。请用中文回答，并且将你的最终答案放在 <Answer> 和 </Answer> 标签之间。在标签内只保留最终答案内容，不要包含任何多余解释或其他文字。 \textit{(Please answer the following question: Who is the host of the program "5 Kids and a Bloke"? Here is some relevant information: "5 Kids and a Bloke" (English: 5 Kids And A Bloke; in Cantonese, meaning "Five Kids and an Uncle") is a reality show produced by Television Broadcasts Limited (TVB), consisting of four episodes. It was broadcast on Jade Channel every Sunday night from October 2 to October 23, 2016 (20:00–20:30) and made available for replay on myTV. The hosts are Tony Hung and the five child actors from the movie "Little Big Master": Ho Yuen-ying, Fu Shun-ying, Wong Sze-nga, Lee Wing-shan, and Chan Lai-yee. Please answer in Chinese and put your final answer between the <Answer> and </Answer> tags. Inside the tags, include only the final answer, without any explanations or extra text.)}\\

answer: 洪永城 \textit{(Tony Hung)}\\
ground truth answer: 是洪永城和电影《五个小孩的校长》中的五个小主角：何涴潆、傅舜盈、王诗雅、李咏珊、陈丽儿。 \textit{(Tony Hung and the five child actors from the movie "Little Big Master": Ho Yuen-ying, Fu Shun-ying, Wong Sze-nga, Lee Wing-shan, and Chan Lai-yee.)}\\
LLMaJ: \textit{NO}\\

\item \textbf{QA answer comparison across different contexts for the same model (Llama3.1-70B): Cantonese-Mandarin}\\
question: 进奏院广东话读音的粤拼是怎样的？ \textit{(What is the Cantonese romanization, of the word} “进奏院” \textit{?)}\\

question-only: \textit{zin3 zau3 jyun2}\\
answer(+standard): \textit{jin4 zhou4 jyun2}\\
answer(+local): \textit{zeon3 zau3 jyun2}\\
answer(+standard+local): \textit{zeon3 zau3 jyun2}\\
answer(+local(translated)): \textit{zeon3 zau3 jyun2}\\
ground truth answer: 进奏院广东话读音的粤拼是zeon3 zau3 jyun2。 \textit{(The Cantonese romanization for} “进奏院” \textit{is zeon3 zau3 jyun2.) }\\

\item \textbf{QA answer comparison across different contexts for the same model (Llama3.1-70B): Bavarian-German}\\
question: \textit{Wo wurde 1983 schon englischer Rap in der Schweiz gespielt? (Where was English rap already played in Switzerland in 1983?)}\\

question-only: \textit{Basel (Basel)}\\
answer(+standard): \textit{Basel (Basel)}\\
answer(+local): \textit{Auf dem Radio-Sender DRS 3. (On the radio station DRS 3.)}\\
answer(+standard+local): \textit{Auf Radio-Senda DRS 3 (On the radio station DRS 3.)}\\
ground truth answer: \textit{1983 wurde auf dem Radiosender DRS 3 schon englischer Rap gespielt. (In 1983, English rap was already played on the radio station DRS 3.)}\\

\item \textbf{QA answer comparison for the same context type (+local) across different models}\\
index: \textit{15}\\
question: 爱尔兰国防军的总司令是谁？真正形式总司令职权的是谁？ \textit{(Who is the Commander-in-Chief of the Irish Defence Forces? Who actually exercises the powers of the Commander-in-Chief?)}\\
prompt: 请回答以下问题：爱尔兰国防军的总司令是谁？真正形式总司令职权的是谁？这里有一些相关的信息：{\CJKfamily{bsmi}愛爾蘭國防軍（愛爾蘭蓋爾文：} \'Oglaigh na h\'Eireann、英文：Irish Defence Forces）{\CJKfamily{bsmi}係愛爾蘭共和國嘅武裝力量，由陸軍、海上部隊、空中部隊同後備國防軍四個部份組成。愛爾蘭國防軍嘅總司令係愛爾蘭總統，但係以總統嘅名義行使總司令職權嘅通常係國防部長，而國防部長嘅諮詢機構係國防委員會。愛爾蘭國防軍個愛爾蘭蓋爾文名，同愛爾蘭共和軍嘅愛爾蘭蓋爾文名完全一樣，反映咗共和軍同國防軍之間上傳下承嘅歷史關係。}请用中文回答，并且将你的最终答案放在 <Answer> 和 </Answer> 标签之间。在标签内只保留最终答案内容，不要包含任何多余解释或其他文字。 
\textit{(Please answer the following question: Who is the Commander-in-Chief of the Irish Defence Forces? Who actually exercises the powers of the Commander-in-Chief? Here is some relevant information: The Irish Defence Forces (Irish: \'Oglaigh na h\'Eireann, English: Irish Defence Forces) are the armed forces of the Republic of Ireland, consisting of the Army, Naval Service, Air Corps, and Reserve Defence Forces. The Commander-in-Chief of the Irish Defence Forces is the President of Ireland, but the powers of the Commander-in-Chief are usually exercised in the President’s name by the Minister for Defence, whose advisory body is the Defence Council. The Irish name of the Defence Forces is identical to that of the Irish Republican Army, reflecting the historical continuity between the IRA and the Defence Forces. Please answer in Chinese and place your final answer between the <Answer> and </Answer> tags. Inside the tags, include only the final answer without any additional explanation or text.)}
\\

llama3.1-8b answer: 爱尔兰总统 \textit{(President of Ireland)}\\
llama3.1-70b answer: {\CJKfamily{bsmi}愛爾蘭總統} \textit{(President of Ireland)}\\
qwen2.5-7b answer: {\CJKfamily{bsmi}愛爾蘭}总统 \textit{(President of Ireland)}\\
qwen2.5-72b answer: {\CJKfamily{bsmi}愛爾蘭總統} \textit{(President of Ireland)}\\
gpt-oss-20b answer: 爱尔兰国防军的总司令是爱尔兰总统；真正行使总司令职权的是国防部长。 \textit{(The Commander-in-Chief of the Irish Defence Forces is the President of Ireland; the one who actually exercises the powers of the Commander-in-Chief is the Minister for Defence.)}\\
gpt-oss-120b answer: 爱尔兰国防军的名义总司令是爱尔兰总统，但实际行使总司令职权的通常是国防部长。 \textit{(The nominal Commander-in-Chief of the Irish Defence Forces is the President of Ireland, but the powers are usually exercised in practice by the Minister for Defence.)}\\
ground truth answer: 爱尔兰国防军的总司令是爱尔兰总统，但是以总统的名义行使总司令职权的通常是国防部长，而国防部长的谘询机构是国防委员会。 \textit{(The Commander-in-Chief of the Irish Defence Forces is the President of Ireland, but the powers of the Commander-in-Chief are usually exercised in the President’s name by the Minister for Defence, whose advisory body is the Defence Council.)}\\
\end{itemize}

\end{CJK*}

\onecolumn % Turns off two-column mode for the rest of the document
\clearpage
\section{Complete results of all metrics}
\label{appx:complete results}

\begin{table}[h!]
\centering
\resizebox{\textwidth}{!}{ 
\begin{tabular}{ll|rrrr|rrrr}
\hline
 \multirow{2}{*}{Model} & \multirow{2}{*}{Context} & \multicolumn{4}{c|}{\yue-\cmn} & \multicolumn{4}{c}{\bay-\deu} \\
\cline{3-10}
 &  & rouge-l & chrF++ & BERTScore & LLMaJ & rouge-l & chrF++ & BERTScore & LLMaJ \\
\hline
\multirow{5}{*}{Llama3.1-8B}   & question-only & 15.66 & 7.02 & 22.42 & 7.50 
& 10.49 & 12.86 & 11.61 & 8.70 \\
% \cline{2-10}
& +standard
&  19.41 &  9.24 &  23.97 &  13.75 
&  13.31 &  14.87 
&  12.79 &  4.35 \\
& +local
& 17.69 & 7.90 & 22.18 & 8.75 
& 11.57 & 13.32 & 11.21 &   4.35 \\
% \cline{2-10}
& +standard+local  &  \textbf{23.16} &  \textbf{10.16} &  \textbf{27.48} & 12.50 
&  \textbf{19.77} &  \textbf{21.11} &  \textbf{19.60} &  \textbf{6.52} \\
& +local (translated)  & 17.96 & 8.33 & 23.48 & 8.75 &
/ & / & / & /
\\
\hline
\multirow{5}{*}{Llama3.1-70B}   &  question-only  
 & 18.85 &  8.05 & 24.48 & 12.50 
 & 15.34 & 17.92 & 22.39 & 15.22
\\
& +standard &
 19.44 & 10.22 & 23.99 & 12.50 
 &   \textbf{24.76} &   \textbf{26.29} &   \textbf{31.00} &   \textbf{21.74} 
\\
& +local & 
   19.69 &   10.69 
&   24.90 &  \textbf{16.25}
& 20.62 & 22.67 & 24.02 & 13.04 
\\
& +standard+local & 
    21.53 & 10.39 &     25.16 &    \textbf{16.25}
& 24.19 & 25.87 & 27.98 & 19.57 
\\
& +local (translated) &
   \textbf{22.08} &    \textbf{10.77} &    \textbf{26.10} &    \textbf{16.25} & 
/ & / & / & /
\\
\hline
\multirow{5}{*}{Qwen2.5-7B}   &  question-only  & 
16.62 & 8.52 & 23.09 &     \textbf{12.50} 
& 21.96 & 27.53 & 27.39 & 0.00 
\\
& +standard 
&  \textbf{20.93} &  \textbf{11.01} 
&  23.89 &  \textbf{12.50} 
&  \textbf{24.68} &  \textbf{31.15} 
&   \textbf{29.82} &   \textbf{6.52}
\\
& +local 
& 19.66 & 8.94 & 22.01 & 11.25 
& 17.32 & 24.45 & 22.22 & 2.17 
\\
& +standard+local & 
20.60 & 9.06 &     \textbf{24.10} & 11.25 
 & 19.50 & 24.50 & 21.27 & 4.35 
\\
& +local (translated) &
17.50 & 8.51 & 18.73 & 7.50 &
/ & / & / & /
\\
\hline
\multirow{5}{*}{Qwen2.5-72B}   &  question-only  & 
19.99 &  9.63  & 25.83  &     17.50  & 
 18.80 & 24.58 & 25.36 & 10.87 
\\
& +standard &
 20.57 & 10.54 & 27.83 & 15.00 
 & 22.69 & 29.87 & 29.38 &     \textbf{15.22}
\\
& +local & 
   21.96 &   10.65 
&   \textbf{29.08} &   16.25 
 &    \textbf{26.58} &    \textbf{33.43} &    \textbf{32.90} & 6.52
\\
& +standard+local 
&     \textbf{22.13} &    \textbf{11.11} & 27.38 &     17.50
 & 26.15 &    \textbf{33.43} & 32.00 &     \textbf{15.22}
\\
& +local (translated) 
& 20.17 & 10.59 & 26.02 &    \textbf{18.75} & 
/ & / & / & /
\\
\hline
\multirow{5}{*}{gpt-oss-20b}   &  question-only  
 & 15.52 &  8.27 & 22.44 & 12.50
 & 15.84 & 20.92 & 22.59 & 8.70 
\\
& +standard &
16.42 &  8.36 & 21.31 & 15.00 
& 18.68 & 23.80 & 26.54 & 8.70 
\\
& +local & 
   18.04 &     9.54 &    23.16 &    16.25
&    \textbf{20.19} &    \textbf{25.28} &    \textbf{28.91} &    \textbf{13.04}
\\
& +standard+local & 
17.80 &     10.17 &    23.68 &    \textbf{18.75}
& 18.30 & 23.70 & 25.68 &    \textbf{13.04}
\\
& +local (translated) &
   \textbf{19.49} &    \textbf{10.40} &    \textbf{26.03} & 15.00 & 
/ & / & / & /
\\
\hline
\multirow{5}{*}{gpt-oss-120b}   &  question-only  & 
15.94 & 9.58 & 23.23 & 15.00 & 
16.47 & 19.74 & 23.38 & 13.04 
\\
& +standard &
    19.47 &    11.25 &    26.28 & 23.75 & 
    21.63 &    26.68 &    29.53 & 15.22 
 \\
& +local & 
18.93 & 11.21 & 24.95 &    26.25 &
20.72 & 26.64 & 29.13 &    \textbf{17.39}
\\
& +standard+local & 
   \textbf{22.58} &    \textbf{12.94} &    \textbf{27.18} &    \textbf{28.75} & 
   \textbf{24.96} &    \textbf{30.93} & 
    \textbf{30.89} & 13.04 
\\
& +local (translated) &
   20.61 &    11.60 & 24.91 & 25.00 & 
/ & / & / & /
\\
\hline
\end{tabular}
}
\caption{Document-level out-of-context QA performance as measured by all metrics. Note that all added contexts lack the information needed to answer the questions.}
\label{tab:results2}
\end{table}

\begin{table}[h!]
\centering
\resizebox{\textwidth}{!}{
\begin{tabular}{ll|rrrr|rrrr}
\hline
\multirow{2}{*}{Model} & \multirow{2}{*}{Context} & \multicolumn{4}{c|}{cmn} & \multicolumn{4}{c}{deu} \\
\cline{3-10}
 &  & rouge-l & chrF++ & BERTScore & LLMaJ & rouge-l & chrF++ & BERTScore & LLMaJ \\
\hline
\multirow{3}{*}{Llama3.1-8b} 
& question-only                 & 17.67 & 12.15 & 27.08 & 10.51 & 9.97 & 16.42 & 29.81 & 13.81 \\
& +context       &  34.27 &  31.38 &    34.80 &   57.36 &    41.87 &    44.13 &    47.69 &    62.46 \\
& +context (translated)     &  \textbf{63.83} &   \textbf{59.88} &   \textbf{63.37} &   \textbf{72.37} &   \textbf{62.07} &    \textbf{64.03} &   \textbf{58.22} &   \textbf{79.88} \\
\hline
\multirow{3}{*}{Llama3.1-70b} 
& question-only                 & 25.49 & 19.29 & 33.79 & 21.62 & 23.44 & 30.08 & 44.94 & 27.33 \\
& +context       &   42.03 &   38.45 &   43.09 &   72.37 &   49.83 &   52.67 &   60.65 &   76.88 \\
& +context (translated)     &   \textbf{74.79} &   \textbf{71.39} &   \textbf{75.52} &   \textbf{86.79} &   \textbf{76.12} &   \textbf{79.15} &   \textbf{79.61} &   \textbf{86.79} \\
\hline
\multirow{3}{*}{Qwen2.5-7b} 
& question-only                 & 21.11 & 13.69 & 30.57 & 12.31 & 8.17  & 15.36 & 23.21 & 7.81 \\
& +context       &   42.56 &    37.82 &    43.91 &    63.66 &    40.46 &    46.51 &    44.22 &    67.27 \\
& +context (translated)     &   \textbf{73.61} &   \textbf{70.82} &   \textbf{74.21} &   \textbf{85.89} &   \textbf{48.33} &   \textbf{56.36} &   \textbf{50.77} &   \textbf{81.38} \\
\hline
\multirow{3}{*}{Qwen2.5-72b} 
& question-only                 & 23.63 & 18.22 & 33.04 & 19.22 & 14.21 & 23.02 & 36.30 & 16.82 \\
& +context       &   48.57 &   44.33 &   51.82 &   72.67 &   46.65 &   54.14 &   51.62 &   80.48 \\
& +context (translated)     &   \textbf{77.09} &   \textbf{73.57} &   \textbf{77.47} &   \textbf{88.59} &   \textbf{63.71} &   \textbf{70.49} &   \textbf{66.37} &   \textbf{90.69} \\
\hline
\multirow{3}{*}{gpt-oss-20b} 
& question-only                 & 15.37 & 13.28 & 24.52 & 15.62 & 10.75 & 18.31 & 32.09 & 15.02 \\
& +context       &   39.84 &   36.24 &   40.16 &   76.88 &   49.92 &   57.16 &   55.16 &   78.68 \\
& +context (translated)     &   \textbf{66.91} &   \textbf{63.63} &   \textbf{67.24} &   \textbf{86.79} &   \textbf{63.11} &   \textbf{69.20} &   \textbf{67.17} &   \textbf{88.29} \\
\hline
\multirow{3}{*}{gpt-oss-120b} 
& question-only                 & 18.99 & 16.23 & 29.38 & 22.52 & 14.23 & 22.69 & 35.38 & 20.72 \\
& +context       &   41.54 &    38.44 &    42.36 &    79.88 &    44.95 &    53.53 &    49.02 &    82.58 \\
& +context (translated)     &    \textbf{65.90} &    \textbf{62.31} &    \textbf{66.44} &    \textbf{87.39} &    \textbf{58.83} &    \textbf{66.05} &    \textbf{61.74} &    \textbf{88.89} \\
\hline
\end{tabular}
}
\caption{QA performance measured by all metrics on the ECLeKTic dataset. Results compare context types of \textit{question-only}, \textit{+context}, and \textit{+context (translated)} across models.}
\label{tab:results_eclektic}
\end{table}

\section{Topic and region results}
\label{appx:all tpoic and region results}
%\section{Full topic-level results}
\begin{table*}[t]
\centering
\resizebox{\textwidth}{!}{
\begin{tabular}{ll|rrrr|rrrr}
\hline
\multirow{2}{*}{Topic} & \multirow{2}{*}{Context} & \multicolumn{4}{c|}{\yue-\cmn} & \multicolumn{4}{c}{\bay-\deu} \\
\cline{3-10}
 &  & rouge-l & chrF++ & BERTScore & LLMaJ & rouge-l & chrF++ & BERTScore & LLMaJ \\
\hline
\multirow{4}{*}{science-math-technology}
& question-only & 19.59 & 11.15 & 22.33 & 22.58 & 10.76 & 19.80 & 13.01 & 33.33 \\
& +standard & 20.82 & 13.34 & 26.42 & 38.71 & 13.47 & 20.25 & 12.71 & 33.33 \\
& +local & 52.86 & 36.85 & 53.71 & 87.10 & 14.05 & 22.69 & 17.89 & 66.67 \\
& +standard+local & 52.74 & 39.43 & 48.60 & 90.32 & 17.28 & 27.27 & 21.12 & 66.67 \\
\hline
\multirow{4}{*}{geography}
& question-only & 15.36 & 11.75 & 25.16 & 17.91 & 19.30 & 24.43 & 28.17 & 9.52 \\
& +standard & 17.86 & 12.71 & 22.99 & 19.40 & 24.36 & 29.37 & 31.08 & 9.52 \\
& +local & 52.50 & 40.49 & 57.02 & 95.52 & 36.24 & 33.23 & 37.23 & 80.95 \\
& +standard+local & 52.54 & 41.04 & 56.34 & 89.55 & 39.79 & 39.68 & 43.67 & 77.78 \\
\hline
\multirow{4}{*}{politics-government}
& question-only & 28.96 & 20.33 & 41.86 & 12.50 & 13.26 & 17.88 & 22.96 & 20.00 \\
& +standard & 32.03 & 24.92 & 35.28 & 25.00 & 19.97 & 21.87 & 24.48 & 30.00 \\
& +local & 59.47 & 50.32 & 66.60 & 75.00 & 28.72 & 30.98 & 35.11 & 80.00 \\
& +standard+local & 59.26 & 48.81 & 63.18 & 62.50 & 30.20 & 35.87 & 41.66 & 60.00 \\
\hline
\multirow{4}{*}{education}
& question-only & 38.95 & 28.78 & 45.86 & 20.00 & 29.06 & 34.44 & 39.73 & 33.33 \\
& +standard & 33.09 & 28.79 & 50.63 & 20.00 & 51.07 & 53.79 & 58.32 & 66.67 \\
& +local & 85.46 & 68.38 & 80.75 & 100.00 & 46.77 & 41.63 & 37.92 & 100.00 \\
& +standard+local & 79.88 & 62.84 & 76.39 & 80.00 & 57.71 & 62.53 & 66.00 & 100.00 \\
\hline
\multirow{4}{*}{transportation}
& question-only & 19.59 & 14.72 & 25.05 & 0.00 & 6.96 & 14.41 & 14.87 & 0.00 \\
& +standard & 24.05 & 18.67 & 29.41 & 8.00 & 12.27 & 26.49 & 18.70 & 0.00 \\
& +local & 62.15 & 49.81 & 65.15 & 96.00 & 13.82 & 23.24 & 14.61 & 66.67 \\
& +standard+local & 64.52 & 53.78 & 63.45 & 84.00 & 21.45 & 20.79 & 16.70 & 100.00 \\
\hline
\multirow{4}{*}{food}
& question-only & 6.50 & 8.01 & 23.13 & 0.00 & 28.65 & 33.24 & 42.23 & 50.00 \\
& +standard & 17.10 & 8.19 & 20.57 & 12.50 & 31.35 & 44.64 & 48.60 & 50.00 \\
& +local & 53.98 & 39.67 & 51.11 & 87.50 & 28.95 & 42.31 & 40.30 & 75.00 \\
& +standard+local & 53.84 & 39.42 & 51.00 & 87.50 & 27.87 & 43.94 & 42.66 & 50.00 \\
\hline
\multirow{4}{*}{linguistics}
& question-only & 15.42 & 4.14 & 20.12 & 0.00 & 11.14 & 16.84 & 22.44 & 16.67 \\
& +standard & 17.04 & 7.01 & 19.84 & 16.67 & 16.99 & 24.82 & 23.37 & 25.00 \\
& +local & 45.30 & 25.83 & 48.70 & 83.33 & 31.12 & 37.85 & 43.94 & 83.33 \\
& +standard+local & 59.60 & 42.43 & 63.90 & 100.00 & 34.99 & 41.05 & 45.74 & 83.33 \\
\hline
\multirow{4}{*}{culture-customs}
& question-only & 11.94 & 6.29 & 10.45 & 12.50 & 31.75 & 34.30 & 41.50 & 41.67 \\
& +standard & 20.49 & 7.71 & 13.62 & 12.50 & 23.61 & 31.18 & 32.85 & 33.33 \\
& +local & 51.93 & 38.79 & 50.74 & 75.00 & 50.32 & 52.24 & 55.31 & 100.00 \\
& +standard+local & 52.15 & 35.92 & 51.79 & 87.50 & 41.80 & 45.13 & 50.05 & 100.00 \\
\hline
\multirow{4}{*}{other}
& question-only & 15.87 & 10.81 & 27.42 & 33.33 & 23.43 & 25.50 & 28.40 & 14.29 \\
& +standard & 12.37 & 8.90 & 25.60 & 33.33 & 31.99 & 40.09 & 38.61 & 28.57 \\
& +local & 40.02 & 26.36 & 36.28 & 66.67 & 37.59 & 44.14 & 47.21 & 85.71 \\
& +standard+local & 42.51 & 26.95 & 38.60 & 66.67 & 39.00 & 51.04 & 55.64 & 85.71 \\
\hline
\multirow{4}{*}{history}
& question-only & 5.45 & 4.40 & 7.59 & 18.18 & 19.34 & 23.53 & 25.55 & 20.83 \\
& +standard & 10.05 & 5.96 & 5.70 & 9.09 & 20.95 & 26.83 & 26.74 & 16.67 \\
& +local & 52.97 & 38.51 & 48.87 & 90.91 & 35.34 & 40.31 & 45.13 & 66.67 \\
& +standard+local & 50.38 & 36.70 & 42.57 & 81.82 & 36.39 & 42.08 & 47.05 & 54.17 \\
\hline
\multirow{4}{*}{entertainment-art}
& question-only & 21.19 & 13.04 & 24.00 & 18.75 & 15.06 & 20.27 & 22.79 & 16.22 \\
& +standard & 25.18 & 14.66 & 28.26 & 16.67 & 20.89 & 25.17 & 27.59 & 24.32 \\
& +local & 56.22 & 39.61 & 56.46 & 83.33 & 33.40 & 36.82 & 40.42 & 86.49 \\
& +standard+local & 57.81 & 42.81 & 57.18 & 79.17 & 35.99 & 40.87 & 44.07 & 83.78 \\
\hline
\end{tabular}
}
\caption{Summary-level QA results as measured by all metrics for \textbf{Qwen2.5-72B} across 11 topics.}
\label{tab:qwen72b_topic_bertscore_complete}
\end{table*}

\begin{table*}[t]
\centering
\resizebox{\textwidth}{!}{ 
\begin{tabular}{ll|rrrr|rrrr}
\hline
\multirow{2}{*}{Region} & \multirow{2}{*}{Context} & \multicolumn{4}{c|}{\yue-\cmn} & \multicolumn{4}{c}{\bay-\deu} \\
\cline{3-10}
 &  & rouge-l & chrF++ & BERTScore & LLMaJ & rouge-l & chrF++ & BERTScore & LLMaJ \\
\hline
\multirow{5}{*}{Local} 
& question-only & 18.81 & 13.03 & 26.04 & 9.90 & 19.01 & 22.62 & 26.86 & 16.67 \\
& +standard & 21.81 & 15.31 & 26.26 & 13.86 & 25.63 & 29.63 & 31.65 & 18.33 \\
& +local & 57.54 & 45.27 & 59.55 & 91.09 & 39.11 & 38.62 & 43.46 & 83.33 \\
& +standard+local & 59.02 & 47.02 & 58.57 & 83.17 & 42.00 & 44.02 & 49.19 & 76.67 \\
& +local (translated) & 60.65 & 47.76 & 59.34 & 89.11 & / & / & / & / \\
\hline
\multirow{5}{*}{Standard} 
& question-only & 17.76 & 12.96 & 23.64 & 13.79 & 20.77 & 29.09 & 32.16 & 27.27 \\
& +standard & 21.04 & 12.77 & 22.82 & 17.24 & 27.97 & 36.88 & 41.35 & 27.27 \\
& +local & 53.59 & 39.55 & 53.13 & 82.76 & 40.62 & 50.52 & 52.68 & 90.91 \\
& +standard+local & 53.41 & 39.09 & 52.23 & 82.76 & 35.17 & 46.90 & 50.19 & 81.82 \\
& +local (translated) & 56.35 & 40.99 & 53.18 & 79.31 & / & / & / & / \\
\hline
\multirow{5}{*}{Other} 
& question-only & 15.83 & 10.03 & 21.33 & 20.73 & 18.08 & 23.24 & 26.48 & 16.82 \\
& +standard & 17.94 & 11.13 & 22.51 & 25.61 & 20.85 & 27.04 & 27.29 & 20.56 \\
& +local & 49.18 & 34.67 & 54.05 & 87.20 & 32.16 & 34.41 & 37.22 & 79.44 \\
& +standard+local & 50.19 & 36.60 & 53.18 & 86.59 & 34.60 & 38.53 & 41.54 & 76.64 \\
& +local (translated) & 56.72 & 39.96 & 54.30 & 85.98 & / & / & / & / \\
\hline
\end{tabular}
}
\caption{Summary-level QA results as measured by all metrics for \textbf{Qwen2.5-72B}, divided by article region classification (Local, Standard, Other).}
\label{tab:qwen72b_region_results_complete}
\end{table*}

\end{document}